\documentclass{article}
\usepackage{ijcai17}

\usepackage{times}
\usepackage{amsmath}
\usepackage{amsfonts}
\usepackage{amssymb}
\usepackage{multirow}
\usepackage{graphicx}
\usepackage{graphics}
\usepackage{epstopdf}
\usepackage[hyperindex,breaklinks]{hyperref}
\title{Compositional Learning of Relation Path Embedding for Knowledge Base Completion}
\author{Xixun Lin\textsuperscript{1}, Yanchun Liang\textsuperscript{1,2}, Fausto Giunchiglia\textsuperscript{3}, Xiaoyue Feng\textsuperscript{1}, Renchu Guan\textsuperscript{1,2}\\
  \textsuperscript{1} Key Laboratory for Symbol Computation and Knowledge Engineering of National Education\\ Ministry, College of Computer Science and Technology, Jilin University, Changchun 130012, China\\
  \textsuperscript{2} Zhuhai Laboratory of Key Laboratory of Symbolic Computation and Knowledge Engineering of \\Ministry of Education, Zhuhai College of Jilin University, Zhuhai 519041, China \\
  \textsuperscript{3} DISI, University of Trento, Italy}

\begin{document}

\maketitle
\begin{abstract}
Large-scale knowledge bases have currently reached impressive sizes;
however, these knowledge bases are still far from complete. In
addition, most of the existing methods for knowledge base completion only
consider the direct links between entities, ignoring the vital impact
of the consistent semantics of relation paths. In this paper, we study
the problem of how to better embed entities and relations of knowledge
bases into different low-dimensional spaces by taking full advantage
of the additional semantics of relation paths, and we propose a
compositional learning model of relation path embedding
(RPE). Specifically, with the corresponding relation and path
projections, RPE can simultaneously embed each entity into two types
of latent spaces. It is also proposed that type constraints could be
extended from traditional relation-specific constraints to the new
proposed path-specific constraints. The results of experiments show that the proposed model achieves significant and consistent improvements compared with the state-of-the-art algorithms.
\end{abstract}

\section{Introduction}
Large-scale knowledge bases (KBs), such as
Freebase~\cite{bollacker2008freebase},
WordNet~\cite{Miller1992WORDNETAL}, Yago~\cite{suchanek2007yago}, and
NELL~\cite{Carlson2010TowardAA}, are critical to natural language
processing applications, e.g., question
answering~\cite{Dong2015QuestionAO}, relation
extraction~\cite{Riedel2013RelationEW}, and language
modeling~\cite{Ahn2016ANK}. These KBs generally contain billions of
facts, and each fact is organized into a triple base format (head
entity, relation, tail entity), abbreviated as
(\emph{h,r,t}). However, the coverage of such KBs is still far from
complete compared with real-world
knowledge~\cite{Dong2014KnowledgeVA}. Traditional KB completion
approaches, such as Markov logic networks
~\cite{Richardson2006MarkovLN}, suffer from feature
sparsity and low efficiency.

\par
Recently, encoding the entire knowledge base into a low-dimensional
vector space to learn latent representations of entity and relation
has attracted widespread attention. These knowledge embedding models
yield better performance in terms of low
complexity and high scalability compared with previous works. Among these methods,
TransE~\cite{Bordes2013TranslatingEF} is a classical neural-based
model, which assumes that each relation can be regarded as a
translation from head to tail and uses a score function
\emph{S}(\emph{h,r,t})=$\|\textbf{h}+\textbf{r}-\textbf{t}\|$ to measure the
plausibility for triples. TransH~\cite{Wang2014KnowledgeGE} and TransR
~\cite{Lin2015LearningEA} are representative variants of TransE. These
variants consider entities from multiple aspects and various relations on different aspects.

\par
 However, the majority of these approaches only exploit direct links
 that connect head and tail entities to predict potential relations
 between entities. These approaches do not explore the fact that
 relation paths, which are denoted as the sequences of relations, i.e.,
 \emph{p}=(\emph{r$_{1}$, r$_{2}$, $\ldots$, r$_{m}$}), play an
 important role in knowledge base completion. For example, the
 sequence of triples (J.K. Rowling, CreatedRole, Harry Potter), (Harry
 Potter, Describedin, Harry Potter and the Philosopher's Stone) can be
 used to infer the new fact (J.K. Rowling, WroteBook, Harry Potter and
 the Philosopher's Stone), which does not appear in the original KBs. Consequently, a promising new research direction is to use relation paths to learn knowledge embeddings~\cite{Neelakantan2015CompositionalVS,Guu2015TraversingKG,Toutanova2016CompositionalLO}.

\par
For a relation path, \emph{consistent semantics} is a semantic
interpretation via composition of the meaning of the component
elements. Each relation path contains its respective consistent
semantics. However, the consistent semantics expressed by some
relation paths \emph{p} is unreliable for reasoning new facts of that
entity pair~\cite{Lin2015ModelingRP}. For instance, there is a common
relation path
$\emph{h}\xrightarrow{HasChildren}\emph{t$^{\prime}$}\xrightarrow{GraduatedFrom}\emph{t}$,
but this path is meaningless for inferring additional relationships between
\emph{h} and \emph{t}. Therefore, reliable relation paths are urgently
needed. Moreover, their consistent semantics, which is essential for
knowledge representation learning, is consistent with the semantics of
relation \emph{r}. Based on this intuition, we propose a compositional
learning model of relation path embedding (RPE), which extends the
projection and type constraints of the specific relation to the specific path.
As the path ranking algorithm
(PRA)~\cite{Lao2011RandomWI} suggests, relation paths that end in many
possible tail entities are more likely to be unreliable for the entity
pair. Reliable relation paths  can thus be filtered using PRA. Figure 1
illustrates the basic idea for relation-specific and path-specific
projections. Each entity is projected by \textbf{M$_{r}$} and
\textbf{M$_{p}$} into the corresponding relation and path spaces. These
different embedding spaces hold the following hypothesis: in the relation-specific space, relation \textbf{r} is regarded as a translation from head \textbf{h$_{r}$} to tail \textbf{t$_{r}$}; likewise, \textbf{p$^{*}$}, the path representation by the composition of relation embeddings, is regarded as a translation from head \textbf{h$_{p}$} to tail \textbf{t$_{p}$} in the path-specific space. We design two types of compositions to dynamically construct the path-specific projection \textbf{M$_{p}$} without extra parameters. Moreover, with slight changes on negative sampling, we also propose that relation-specific and path-specific type constraints can be seamlessly incorporated into our model.
 \par
Our main contributions are as follows: \par
1) To reinforce the reasoning ability of knowledge embedding models, the consistent semantics and the path spaces are introduced. \par
2) The path-specific type constraints generated from path space can help to improve the model's discriminability. \par
3) Compared with the pure data-driven mechanism in the knowledge
embedding models used, the way in which we utilize PRA to find reliable relation paths improves the knowledge representation learning interpretability.
\par
The remainder of this paper is organized as follows. We first provide
a brief review of related knowledge embedding models in Section 2. The
details of RPE are introduced in Section 3. The experiments and
analysis are reported in Section 4. Conclusions and directions for
future work are reported in the final section.

\begin{figure}
\begin{center}
\includegraphics[width=3.2in,height=2.5in]{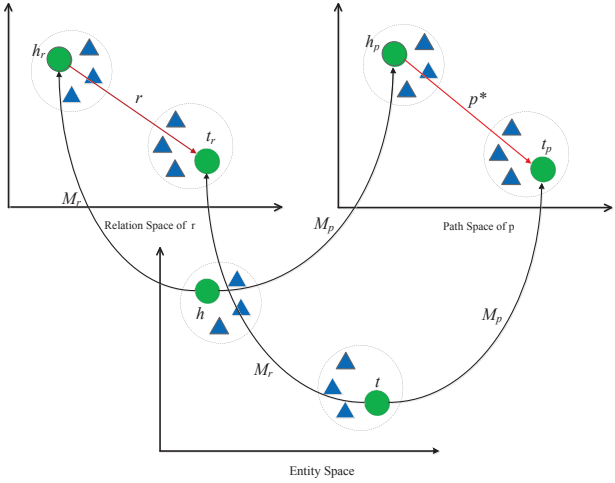}
\caption{Simple illustration of relation-specific and path-specific projections.}
\label{RPE_fig1}
\end{center}
\end{figure}

\section{Related Work}
We first concentrate on three classical translation-based models that
only consider direct links between entities. The bold lowercase letter
\textbf{v} denotes a column vector, and the bold uppercase letter
\textbf{M} denotes a matrix. The first translation-based model is
TransE, and it defines the score function as
\emph{S}(\emph{h,r,t})=$\|\textbf{h}+\textbf{r}-\textbf{t}\|$ for each triple
(\emph{h,r,t}). The score will become smaller if the triple
(\emph{h,r,t}) is correct; otherwise, the score will become higher. The
embeddings are learned by optimizing a global margin-loss
function. This assumption is clearly simple and cannot address more
complex relation attributes well, i.e., 1-to-N, N-to-1, and N-to-N. To
alleviate this problem, TransH projects entities into a
relation-dependent hyperplane by the normal vector \textbf{w$_{r}$}:
\textbf{h$_{h}$}=\textbf{h}-\textbf{w$_r^T$hw$_r$} and
\textbf{t$_{h}$}=\textbf{t}-\textbf{w$_r^T$tw$_r$} (restrict
$\|$\textbf{w$_{r}$}$\|$$_{2}$=1). The corresponding score function is
\emph{S}(\emph{h,r,t})=$\|\textbf{h$_{h}$}+\textbf{r}-\textbf{t$_{h}$}\|$. TransE
and TransH achieve translations on the same embedding space, whereas
TransR assumes that each relation should be used to project entities
into different relation-specific embedding spaces since different
relations may place emphasis on different entity aspects. The
projected entity vectors are \textbf{h$_{r}$=M$_{r}$h} and
\textbf{t$_{r}$=M$_{r}$t}; thus, the new score function is defined as \emph{S}(\emph{h,r,t})=$\|\textbf{h$_{r}$}+\textbf{r}-\textbf{t$_{r}$}\|$.

\par
Another research direction focuses on improving the prediction
performance by using prior knowledge in the form of
relation-specific type
constraints~\cite{Krompass2015TypeConstrainedRL,Chang2014TypedTD,Wang2015KnowledgeBC}. Note
that each relation should possess \emph{Domain} and \emph{Range}
fields to indicate the subject and object types, respectively. For
example, the relation haschildren's \emph{Domain} and \emph{Range} types both belong to a person. By exploiting these limited rules, the harmful influence of a merely data-driven pattern can be avoided. Type-constrained TransE~\cite{Krompass2015TypeConstrainedRL} imposes these constraints on the global margin-loss function to better distinguish similar embeddings in latent space.
\par
A third current related work is PTransE~\cite{Lin2015ModelingRP} and
the path ranking algorithm (PRA)~\cite{Lao2011RandomWI}. PTransE
considers relation paths as translations between head and tail
entities and primarily addresses two problems: 1) exploit a variant of
PRA to select reliable relation paths, and 2) explore three path
representations by compositions of relation embeddings. PRA, as one of
the most promising research innovations for knowledge base completion,
has also attracted considerable
attention~\cite{Lao2015LearningRF,Gardner2015EfficientAE,Wang2016KnowledgeBC,Nickel2016ARO}. PRA
uses the path-constrained random walk probabilities as path features
to train linear classifiers for different relations. In large-scale
KBs, relation paths have great significance for enhancing the
reasoning ability for more complicated situations. However, none of the
aforementioned models take full advantage of the consistent semantics of relation paths.
\section{Our Model}
 The consistent semantics expressed by reliable relation paths has a
 significant impact on learning meaningful embeddings. Here, we
 propose a compositional learning model of relation path embedding
 (RPE), which includes novel path-specific projection and type
 constraints. All entities constitute
 the entity set $\zeta$, and all relations constitute the relation set
 \emph{R}. RPE uses PRA to select reliable relation paths. Precisely,
 for a triple (\emph{h,r,t}),
 P$_{all}$=\{\emph{p$_{1}$,p$_{2}$,$\ldots$,p$_{k}$}\} is the path set
 for the entity pair (\emph{h,t}).  PRA calculates
 \emph{P}(\emph{t}$\vert$\emph{h}, \emph{p$_{i}$}), the probability of
 reaching \emph{t} from \emph{h} following the sequence of relations
 indicated by \emph{p$_{i}$}, which can be recursively defined as follows: \\If \emph{p$_{i}$} is an empty path,
\begin{equation}
   P(\emph{t}\vert\emph{h},\emph{p$_{i}$})=
   \left\{
   \begin{aligned}
   &1 \quad if \ \emph{h}=\emph{t} \\
   &0 \quad otherwise
   \end{aligned}
   \right.
  \end{equation}
If \emph{p$_{i}$} is not an empty path, then \emph{p$^{\prime}_{i}$} is defined as \emph{r$_{1}$,$\ldots$,r$_{m-1}$}; subsequently,
\begin{equation}
P(\emph{t}\vert\emph{h},\emph{p$_{i}$})=\sum_{{t^{\prime}}\in\emph{Ran(p$^{\prime}_{i}$)}}P(\emph{t$^{\prime}$}\vert\emph{h},\emph{p$^{\prime}_{i}$})\cdot P(\emph{t}\vert\emph{t$^{\prime}$},\emph{r$_{m}$})
\end{equation}
\emph{Ran(p$^{\prime}_{i}$)} is the set of ending nodes of p$^{\prime}_{i}$. RPE obtains the reliable relation paths set P$_{filter}$=\{\emph{p$_{1}$,p$_{2}$,$\ldots$,p$_{z}$}\} by selecting relation paths whose probabilities are above a certain threshold $\eta$.
\subsection{Path-specific Projection}
The key idea of RPE is that the consistent semantics of reliable
relation paths is similar to the semantics of the relation between an
entity pair. For a triple (\emph{h,r,t}), RPE exploits projection
matrices \textbf{M$_{r}$}, \textbf{M$_{p}$} $\in$ $\mathbb{R}^{m\times
  n}$ to project entity vectors \textbf{h}, \textbf{t} $\in$
$\mathbb{R}^{n}$ in entity space into the corresponding relation and
path spaces simultaneously (\emph{m} is the dimension of relation embeddings,
\emph{n} is the dimension of entity embeddings, and \emph{m} may differ from \emph{n}). The
projected vectors (\textbf{h$_{r}$}, \textbf{h$_{p}$},
\textbf{t$_{r}$}, \textbf{t$_{p}$}) in their respective embedding spaces are denoted as follows:
\begin{equation}
\begin{split}
\textbf{h$_{r}$}=\textbf{M$_{r}$}\textbf{h},\qquad \textbf{h$_{p}$}=\textbf{M$_{p}$}\textbf{h}
\end{split}
\end{equation}
\begin{equation}
\begin{split}
\textbf{t$_{r}$}=\textbf{M$_{r}$}\textbf{t},\qquad \textbf{t$_{p}$}=\textbf{M$_{p}$}\textbf{t}
\end{split}
\end{equation}
Because relation paths are those sequences of relations
\emph{p}=(\emph{r$_{1}$, r$_{2}$, $\ldots$, r$_{m}$}), we dynamically
use \textbf{M$_{r}$} to construct \textbf{M$_{p}$} to decrease the model complexity. Subsequently, we explore two
compositions for the formation of \textbf{M$_{p}$}, which are
formulated as follows:
\begin{equation}
\begin{split}
\textbf{M$_{p}$}=&\textbf{M$_{r_{1}}$}+\textbf{M$_{r_{2}}$}+\ldots+\textbf{M$_{r_{m}}$} \\ &(addition~composition)
\end{split}
\end{equation}
\begin{equation}
\begin{split}
\textbf{M$_{p}$}=&\textbf{M$_{r_{1}}$}\times\textbf{M$_{r_{2}}$}\times\ldots\times\textbf{M$_{r_{m}}$}\\ &(multiplication~composition)
\end{split}
\end{equation}
where addition composition (ACOM) and multiplication composition
(MCOM) represent cumulative addition and multiplication for path
projection. Matrix normalization is applied on \textbf{M$_{p}$} for
both compositions. The new score function is defined as follows:
\begin{equation}
\begin{split}
&G(\emph{h,r,t}) =S(\emph{h,r,t})+\lambda \cdot S(\emph{h,p,t})=\|\textbf{h$_{r}$}+\textbf{r}-\textbf{t$_{r}$}\|+ \\
&\frac{\lambda}{\emph{Z}}\sum_{\emph{p$_{i}$}\in\emph{P$_{filter}$}}P(\emph{t}\vert\emph{h},\emph{p$_{i}$})\cdot P_{\emph{r}}(\emph{r}\vert\emph{p$_{i}$})\cdot \|\textbf{h$_{p}$}+\textbf{p$_{i}^{*}$}-\textbf{t$_{p}$}\|
\end{split}
\end{equation}
For path representation \textbf{p$^{*}$}, we use
\textbf{p$^{*}$}=\textbf{r$_{1}$}+\textbf{r$_{2}$}+\ldots+\textbf{r$_{m}$},
as suggested by PTransE. $\lambda$ is the hyper-parameter used to
balance the knowledge embedding score \emph{S}(\emph{h,r,t}) and the relation
path embedding score \emph{S}(\emph{h,p,t}). Z=$\sum_{p_{i}\in
  p_{filter}}$\emph{P}(\emph{t}$\vert$\emph{h}, \emph{p$_{i}$}) is the
normalization factor, and
\emph{P$_{r}$}(\emph{r}$\vert$\emph{p$_{i}$}) =
\emph{P$_{r}$}(\emph{r}, \emph{p$_{i}$}) /
\emph{P$_{r}$}(\emph{p$_{i}$}) is used to assist in calculating the
reliability of relation paths. In the experiments, we increase the
limitation on these embeddings, i.e., $\|\textbf{h}\|_{2}$ $\leqslant$
1,  $\|\textbf{t}\|_{2}$ $\leqslant$ 1,  $\|\textbf{r}\|_{2}$
$\leqslant$ 1, $\|\textbf{h$_{r}$}\|_{2}$ $\leqslant$ 1,
$\|\textbf{t$_{r}$}\|_{2}$ $\leqslant$ 1, $\|\textbf{h$_{p}$}\|_{2}$
$\leqslant$ 1, and $\|\textbf{t$_{p}$}\|_{2}$ $\leqslant$ 1. By exploiting the consistent semantics of reliable relation paths, RPE embeds entities into the relation and path spaces simultaneously. This method improves the flexibility of RPE when modeling more complicated relation attributes.

\subsection{Path-specific Type Constraints}
In RPE, based on the semantic similarity between relations and reliable relation paths,
we can naturally extend the relation-specific type constraints to novel path-specific type constraints. In type-constrained TransE, the distribution of corrupted triples is a uniform distribution.

\par
In our model, we borrow the idea from TransH, incorporating the two
type constraints with a Bernoulli distribution. For each relation
\emph{r}, we denote the \emph{Domain$_{r}$} and \emph{Range$_{r}$} to
indicate the subject and object types of relation
\emph{r}. $\zeta_{Domain_{r}}$ is the entity set whose entities
conform to \emph{Domain$_{r}$}, and $\zeta_{Range_{r}}$ is the entity
set whose entities conform to \emph{Range$_{r}$}. We calculate the
average numbers of tail entities for each head entity, named
\emph{teh}, and the average numbers of head entities for each tail
entity, named \emph{het}. The Bernoulli distribution with parameter
$\frac{teh}{teh+het}$ for each relation \emph{r} is incorporated with
the two type constraints, which can be defined as follows: RPE samples
entities from $\zeta_{Domain_{r}}$ to replace the head entity with
probability $\frac{teh}{teh+het}$, and it samples entities from
$\zeta_{Range_{r}}$ to replace the tail entity with probability
$\frac{het}{teh+het}$. The objective function for
RPE is defined as follows:
\begin{equation}
\begin{split}
L~=~ \sum_{\emph{(h,r,t)}\in \emph{C}}&\big[L(h,r,t)+\frac{\lambda}{Z}\sum_{p_{i}\in P_{filter}}P(\emph{t}\vert\emph{h},\emph{p$_{i}$})\cdot \\
                                      &P_{\emph{r}}(\emph{r}\vert\emph{p$_{i}$})L(h,p_{i},t)\big]
\end{split}
\end{equation}
\emph{L}(\emph{h,r,t}) is the loss function for triples, and \emph{L}(\emph{h,$p_{i}$,t}) is the loss function for relation paths.
\begin{equation}
\begin{split}
L(h,r,t)=\sum_{(h^{\prime},r,t^{\prime})\in C^{\prime\prime}}&max(0,S(h,r,t)+\gamma_{1}-\\&S(h^{\prime},r,t^{\prime}))
\end{split}
\end{equation}
\begin{equation}
\begin{split}
L(h,p_{i},t)=\sum_{(h^{\prime},r,t^{\prime})\in C^{\prime\prime}}&max(0,S(h,p_{i},t)+\gamma_{2}-\\&S(h^{\prime},p_{i},t^{\prime}))
\end{split}
\end{equation}
We denote \emph{C}=\{(\emph{h${_i}$},\emph{r${_i}$},\emph{t${_i}$})
$\vert$ \emph{i=1,2$\ldots$,t}\} as the set of all observed triples
and
\emph{C$^{\prime}$}=\{(\emph{h${_{i}^{\prime}}$},\emph{r${_i}$},\emph{t${_i}$})
$\cup$ \emph{(h$_{i}$},\emph{r$_{i}$},\emph{t$_{i}^{\prime}$}) $\vert$
\emph{i=1,2$\ldots$,t}\} as the set of corrupted triples, where each
element of \emph{C$^{\prime}$} is obtained by randomly sampling from
$\zeta$. \emph{C$^{\prime\prime}$}, whose element conforms to the two
type constraints with a Bernoulli distribution, is a subset of
\emph{C$^{\prime}$}. The \emph{Max}(\emph{0, x}) returns the maximum
between \emph{0} and \emph{x}. $\gamma$ is the hyper-parameter of
margin, which separates corrected triples and corrupted triples. By
exploiting these two prior knowledges, RPE could better distinguish
similar embeddings in different embedding spaces, thus allowing it to achieve better prediction.

\subsection{Training Details}
We adopt stochastic gradient descent (SGD) to minimize the objective function. TransE or RPE (initial) can be exploited for the initializations of all entities and relations. The score function of RPE (initial) is as follows:
\begin{equation}
\begin{split}
&G(\emph{h,r,t}) =S(\emph{h,r,t})+\lambda\cdot S(\emph{h,p,t})=\|\textbf{h}+\textbf{r}-\textbf{t}\|+ \\
&\frac{\lambda}{\emph{Z}}\sum_{\emph{p$_{i}$}\in\emph{P$_{filter}$}}P(\emph{t}\vert\emph{h},\emph{p$_{i}$})\cdot P_{\emph{r}}(\emph{r}\vert\emph{p$_{i}$})\cdot \|\textbf{h}+\textbf{p$_{i}^{*}$}-\textbf{t}\|
\end{split}
\end{equation}
We also employ this score function in our experiment as a baseline. The projection matrices \textbf{M}s are initialized as identity matrices. RPE holds the local closed-world assumption (LCWA)~\cite{Dong2014KnowledgeVA}, where each relation's domain and range types are based on the instance level. Their type information is provided by KBs or the entities shown in observed triples.

\par
Note that each relation \emph{r} has the reverse relation
\emph{$r^{-1}$}; therefore, to increase supplemental path information,
RPE utilizes the reverse relation paths. For
example, for the relation path $\emph{LeBron James}
\xrightarrow{PlayFor} \emph{Cleveland Cavaliers}
\xrightarrow{BelongTo} \emph{NBA}$, its reverse relation path can be defined as $\emph{NBA} \xrightarrow{BelongTo^{-1}} \emph{Cleveland Cavaliers} \xrightarrow{PlayFor^{-1}} \emph{LeBron James}$.

\par
For each iteration, we randomly sample a correct triple (\emph{h,r,t})
with its reverse (\emph{t,r$^{-1}$,h}), and the final score function
of our model is defined as follows:
\begin{equation}
\begin{split}
F(\emph{h,r,t})=G(\emph{h,r,t})+G(\emph{t,r$^{-1}$,h})
\end{split}
\end{equation}
Theoretically, we can arbitrarily set the length of the relation path,
but in the implementation, we prefer to take a smaller value to reduce
the time required to enumerate all relation paths. Moreover, as
suggested by the path-constrained random walk probability
\emph{P}(\emph{t$\vert$h, p}), as the path length increases, \emph{P}(\emph{t$\vert$h, p}) will become smaller and the relation path will more likely be cast off.
\section{Experiments}
To examine the retrieval and discriminative ability of our model, RPE is evaluated on two standard subtasks of knowledge base completion: link prediction and triple classification. We also present a case study on consistent semantics learned by our method to further highlight the importance of relation paths for knowledge representation learning.
\subsection{Datasets}
We evaluate our model on two classical large-scale knowledge bases:
Freebase and WordNet. Freebase is a large collaborative knowledge base
that contains billions of facts about the real world, such as the
triple (Beijing, Locatedin, China), which describes the fact that
Beijing is located in China. WordNet is a large lexical knowledge base
of English, in which each entity is a synset that expresses a distinct
concept, and each relationship is a conceptual-semantic or lexical
relation. We use two subsets of Freebase, FB15K and
FB13~\cite{Bordes2013TranslatingEF}, and one subset of WordNet,
WN11~\cite{Socher2013ReasoningWN}. Table 1 presents the statistics of
the datasets, where each column represents the numbers of entity type, relation type and triples that have been split into training, validation and test sets.
\begin{table}
\setlength{\abovecaptionskip}{0pt}
\setlength{\belowcaptionskip}{5pt}
\centering
\caption{The statistics of the datasets.}
\resizebox{7.5cm}{!}{
\begin{tabular}{crrrrr}
\hline
Dataset &\#Ent &\#Rel &\#Train &\#Valid &\#Test\\ \hline  % \hline 在此行下面画一横线
FB15K &14591 &1345 &483142 &50000 &59071\\         % \\ 表示重新开始一行
FB13 &75043 &13 &316232 &5908 &23733\\        % & 表示列的分隔线
WN11 &38696 &11 &112581 &2609 &10544\\ \hline
\end{tabular}
}
\end{table}
\begin{table}
\setlength{\abovecaptionskip}{0pt}
\setlength{\belowcaptionskip}{5pt}
\centering
\caption{Evaluation results on link prediction.}
\resizebox{8.5cm}{!}{
\begin{tabular}{|cccc|cc|}
\hline
\multicolumn{2}{|c|}{\multirow{2}{*}{Metric}} &\multicolumn{2}{c|}{Mean Rank} &\multicolumn{2}{c|}{Hits@10(\%)} \\
\multicolumn{2}{|c|}{}                        &Raw  &Filter                    &Raw   &Filter\\ \hline     % \\ 表示重新开始一行
\multicolumn{2}{|c|}{TransE~\cite{Bordes2013TranslatingEF}}                  &243  &125                       &34.9  &47.1 \\        % & 表示列的分隔线
\multicolumn{2}{|c|}{TransH (unif)~\cite{Wang2014KnowledgeGE}}            &211  &84                        &42.5  &58.5 \\
\multicolumn{2}{|c|}{TransH (bern)~\cite{Wang2014KnowledgeGE}}            &212  &87                        &45.7  &64.4 \\
\multicolumn{2}{|c|}{TransR (unif)~\cite{Lin2015LearningEA}}            &226  &78                        &43.8  &65.5 \\
\multicolumn{2}{|c|}{TransR (bern)~\cite{Lin2015LearningEA}}            &198  &77                        &48.2  &68.7 \\
\multicolumn{2}{|c|}{PTransE (ADD, 2-hop)~\cite{Lin2015ModelingRP}}                 &200  &54                        &51.8  &83.4 \\
\multicolumn{2}{|c|}{PTransE (MUL, 2-hop)~\cite{Lin2015ModelingRP}}                 &216  &67                        &47.4  &77.7 \\
\multicolumn{2}{|c|}{PTransE (ADD, 3-hop)~\cite{Lin2015ModelingRP}}                 &207  &58                        &51.4  &84.6 \\
\multicolumn{2}{|c|}{TranSparse (separate, S, unif)~\cite{Ji2016KnowledgeGC}}       &211  &63                        &50.1  &77.9 \\
\multicolumn{2}{|c|}{TranSparse (separate, S, bern)~\cite{Ji2016KnowledgeGC}}       &187  &82                        &\textbf{53.3}  &79.5 \\
\hline
\multicolumn{2}{|c|}{RPE (initial)}           &207  &58                       &50.8  &82.2 \\
\multicolumn{2}{|c|}{RPE (PC)}  &196  &77                       &49.1  &72.6 \\
\multicolumn{2}{|c|}{RPE (ACOM)}            &\textbf{171}&\textbf{41}       &52.0  &\textbf{85.5}\\
\multicolumn{2}{|c|}{RPE (MCOM)}            &183  &43                       &52.2  &81.7 \\
\multicolumn{2}{|c|}{RPE (PC + ACOM)}            &184  &42      &51.1  &84.2 \\
\multicolumn{2}{|c|}{RPE (PC + MCOM)}            &186  &43      &51.7  &76.5 \\
\hline
\end{tabular}
}
\end{table}
\par
In our model, each triple has its own reverse triple for increasing
the reverse relation paths. Therefore, the total number of
triples is twice the number used in the original datasets. Our model
exploits the LCWA. In this case, we utilize the type information
provided by~\cite{Xie2016RepresentationLO} for FB15K. As for FB13 and
WN11, we do not depend on the auxiliary data, and the domain and range of each relation are approximated by the triples from
the original datasets.
\begin{table*}
\setlength{\abovecaptionskip}{0pt}
\setlength{\belowcaptionskip}{5pt}
\centering
\caption{Evaluation results on FB15K by mapping properties of relations (\%).}
\resizebox{16cm}{!}{
\begin{tabular}{|cc|cccc|cccc|}
\hline
\multicolumn{2}{|c|}{Tasks}
&\multicolumn{4}{|c|}{Predicting Head Entities (Hits@10)}
&\multicolumn{4}{|c|}{Predicting Tail Entities (Hits@10)} \\
\hline
\multicolumn{2}{|c|}{Relation Category}  &1-to-1 &1-to-N &N-to-1 &N-to-N                             &1-to-1 &1-to-N &N-to-1 &N-to-N \\
\hline
\multicolumn{2}{|c|}{TransE~\cite{Bordes2013TranslatingEF}}             &43.7   &65.7   &18.2   &47.2                               &43.7   &19.7   &66.7   &50.0 \\
\multicolumn{2}{|c|}{TransH (unif)~\cite{Wang2014KnowledgeGE}}             &66.7   &81.7   &30.2   &57.4                         &63.7   &30.1   &83.2   &60.8 \\
\multicolumn{2}{|c|}{TransH (bern)~\cite{Wang2014KnowledgeGE}}             &66.8   &87.6   &28.7   &64.5                         &65.5   &39.8   &83.3   &67.2 \\
\multicolumn{2}{|c|}{TransR (unif)~\cite{Lin2015LearningEA}}             &76.9   &77.9   &38.1   &66.9                         &76.2   &38.4   &76.2   &69.1 \\
\multicolumn{2}{|c|}{TransR (bern)~\cite{Lin2015LearningEA}}             &78.8   &89.2   &34.1   &69.2                         &79.2   &37.4   &90.4   &72.1 \\
\multicolumn{2}{|c|}{PTransE (ADD, 2-hop)~\cite{Lin2015ModelingRP}}             &91.0   &92.8   &60.9   &83.8                         &91.2   &74.0   &88.9   &86.4 \\
\multicolumn{2}{|c|}{PTransE (MUL, 2-hop)~\cite{Lin2015ModelingRP}}             &89.0   &86.8   &57.6   &79.8                         &87.8   &71.4   &72.2   &80.4 \\
\multicolumn{2}{|c|}{PTransE (ADD, 3-hop)~\cite{Lin2015ModelingRP}}             &90.1   &92.0   &58.7   &86.1                         &90.7   &70.7   &87.5   &88.7 \\
\multicolumn{2}{|c|}{TranSparse (separate, S, unif)~\cite{Ji2016KnowledgeGC}}   &82.3   &85.2   &51.3   &79.6                         &82.3   &59.8   &84.9   &82.1 \\
\multicolumn{2}{|c|}{TranSparse (separate, S, bern)~\cite{Ji2016KnowledgeGC}}   &86.8   &95.5   &44.3   &80.9                         &86.6   &56.6   &94.4   &83.3 \\
\hline
\multicolumn{2}{|c|}{RPE (initial)}             &83.9   &93.6   &60.1   &78.2                         &82.2   &66.8   &92.2   &80.6 \\
\multicolumn{2}{|c|}{RPE (PC)}         &82.6   &92.7   &44.0   &71.2                         &82.6   &64.6   &81.2   &75.8 \\
\multicolumn{2}{|c|}{RPE (ACOM)}          &\textbf{92.5}   &\textbf{96.6}   &\textbf{63.7}   &\textbf{87.9}    &\textbf{92.5}  &\textbf{79.1}  &\textbf{95.1}   &\textbf{90.8} \\
\multicolumn{2}{|c|}{RPE (MCOM)}          &91.2   &95.8   &55.4   &87.2                         &91.2   &66.3   &94.2   &89.9 \\
\multicolumn{2}{|c|}{RPE (PC + ACOM)}             &89.5   &94.3   &63.2   &84.2                 &89.1   &77.0   &89.7   &87.6 \\
\multicolumn{2}{|c|}{RPE (PC + MCOM)}             &89.3   &95.6   &45.2   &84.2                 &89.7   &62.8   &94.1   &87.7 \\
\hline
\end{tabular}
}
\end{table*}
\subsection{Link Prediction}
The link prediction task consists of predicting the possible \emph{h} or \emph{t} for test triples when \emph{h} or \emph{t} is missed. FB15K is employed for this task.

\subsubsection{Evaluation Protocol} We follow the same evaluation
procedures as used
in~\cite{Bordes2013TranslatingEF,Wang2014KnowledgeGE,Lin2015LearningEA}. First,
for each test triple (\emph{h,r,t}), we replace \emph{h} or \emph{t}
with every entity in $\zeta$. Second, each corrupted triple is
calculated by the corresponding score function \emph{S}(\emph{h,r,t}). The final step is to rank the original correct entity with these scores in descending order.

\par
Two evaluation metrics are reported: the average rank of correct
entities (Mean Rank) and the proportion of correct entities ranked in
the top 10 (Hits@10). Note that if a corrupted triple already exists
in the knowledge base, then it should not be considered to be
incorrect. We prefer to remove these corrupted triples from our
dataset, and call this setting "Filter". If these corrupted triples
are reserved, then we call this setting "Raw". In both settings, if
the latent representations of entity and relation are better, then a lower mean rank and a higher Hits@10 should be achieved. Because we use the same dataset, the baseline results reported in~\cite{Lin2015LearningEA,Lin2015ModelingRP,Ji2016KnowledgeGC} are directly used for comparison.

\subsubsection{Implementation}
 We set the dimension of entity embedding \emph{m} and relation embedding \emph{n} among \{20, 50, 100, 120\}, the margin $\gamma_{1}$ among \{1, 2, 3, 4, 5\}, the margin $\gamma_{2}$ among \{3, 4, 5, 6, 7, 8\}, the learning rate $\alpha$ for SGD among \{0.01, 0.005, 0.0025, 0.001, 0.0001\}, the batch size \emph{B} among \{20, 120, 480, 960, 1440, 4800\}, and the balance factor $\lambda$ among \{0.5, 0.8, 1,1.5, 2\}. The threshold $\eta$ was set in the range of \{0.01, 0.02, 0.04, 0.05\} to reduce the calculation of meaningless paths.

\par
Grid search is used to determine the optimal parameters. The best
configurations for RPE (ACOM) are \emph{n}=100, \emph{m}=100,
$\gamma_{1}$=2, $\gamma_{2}$=5, $\alpha$=0.0001, \emph{B}=4800,
$\lambda$=1, and $\eta$=0.05. We select RPE (initial) to initialize
our model, set the path length as 2, take \emph{L$_{1}$} norm for the score function, and traverse our model for 500 epochs.

\subsubsection{Analysis of Results}
Table 2 reports the results of link prediction, in which the first
column is translation-based models, the variants of PTransE, and our
models. The numbers in bold are the best performance, and n-hop
indicates the path length \emph{n} that PTransE exploits. We denote
RPE only with path-specific constraints as RPE (PC), and from the
results, we can observe the following. 1) Our models significantly
outperform the classical knowledge embedding models (TransE, TransH,
TransR, and TranSparse) and PTransE on FB15K with the metrics of mean
rank and Hits@10 (filter). The results demonstrate that the
path-specific projection can explore further
implications on relation paths, which are crucial for knowledge base
completion. 2) RPE (PC) improves slightly compared with the
baselines. We believe that this result is primarily because RPE (PC)
only focuses on local information provided by related embeddings,
ignoring some global information compared with the approach of
randomly selecting corrupted entities. In terms of mean rank, RPE
(ACOM) achieves the best performance with 14.5\% and 24.1\% error
reduction compared with PTransE's performance in the raw and filter
settings, respectively. In terms of Hits@10, RPE (ACOM) brings few improvements. RPE with path-specific type constraints and projection (RPE (PC + ACOM) and RPE (PC + MCOM)) is a compromise between RPE (PC) and RPE (ACOM).
\par
Table 3 presents the separated evaluation results by mapping
properties of relations on
FB15K. Mapping properties of relations follows the same rules in~\cite{Bordes2013TranslatingEF},
and the metrics are Hits@10 on head and tail entities. From Table
3, we can conclude that 1) RPE (ACOM) outperforms all baselines in all
mapping properties of relations. In particular, for the 1-to-N,
N-to-1, and N-to-N types of relations that plague knowledge embedding
models, RPE (ACOM) improves 4.1\%, 4.6\%, and 4.9\% on head entity's
prediction and 6.9\%, 7.0\%, and 5.1\% on tail entity's prediction
compared with previous state-of-the-art performances achieved by
PTransE (ADD, 2-hop). 2) RPE (MCOM) does not perform as well as RPE
(ACOM), and we believe that this result is because RPE's path representation is not consistent with RPE (MCOM)'s composition of projections. Although RPE (PC) improves little compared with PTransE, we will indicate the effectiveness of relation-specific and path-specific type constraints in triple classification. 3) We use the relation-specific projection to construct path-specific ones dynamically; then, entities are encoded into relation-specific and path-specific spaces simultaneously. The experiments are similar to link prediction, and the results of experiments results further demonstrate the better expressibility of our model.
\begin{table}
\setlength{\abovecaptionskip}{0pt}
\setlength{\belowcaptionskip}{5pt}
\centering
\caption{Evaluation results of triple classification (\%).}
\resizebox{8.5cm}{!}{
\begin{tabular}{|cc|c|c|c|}
\hline
\multicolumn{2}{|c|}{Datasets}       &WN11 &FB13 &FB15K \\
\hline
\multicolumn{2}{|c|}{TransE (unif)~\cite{Bordes2013TranslatingEF}}   &75.9  &70.9 &77.8  \\    % \\ 表示重新开始一行
\multicolumn{2}{|c|}{TransE (bern)~\cite{Bordes2013TranslatingEF}}   &75.9  &81.5 &85.3  \\
\multicolumn{2}{|c|}{TransH (unif)~\cite{Wang2014KnowledgeGE}}   &77.7  &76.5 &78.4  \\
\multicolumn{2}{|c|}{TransH (bern)~\cite{Wang2014KnowledgeGE}}   &78.8  &83.3 &85.8  \\
\multicolumn{2}{|c|}{TransR (unif)~\cite{Lin2015LearningEA}}   &85.5  &74.7 &79.2  \\
\multicolumn{2}{|c|}{TransR (bern)~\cite{Lin2015LearningEA}}   &85.9  &82.5 &87.0  \\
\multicolumn{2}{|c|}{PTransE (ADD, 2-hop)~\cite{Lin2015ModelingRP}}        &80.9  &73.5 &83.4  \\
\multicolumn{2}{|c|}{PTransE (MUL, 2-hop)~\cite{Lin2015ModelingRP}}        &79.4  &73.6 &79.3 \\
\multicolumn{2}{|c|}{PTransE (ADD, 3-hop)~\cite{Lin2015ModelingRP}}        &80.7  &73.3 &82.9  \\
\hline
\multicolumn{2}{|c|}{RPE (initial)}  &80.2  &73.0 &68.8  \\
\multicolumn{2}{|c|}{RPE (PC)}       &83.8  &77.4 &77.9  \\
\multicolumn{2}{|c|}{RPE (ACOM)}   &84.7  &80.9 &85.4  \\
\multicolumn{2}{|c|}{RPE (MCOM)}   &83.6  &76.2 &85.1  \\
\multicolumn{2}{|c|}{RPE (PC + ACOM)}  &\textbf{86.8}  &\textbf{84.3}  &\textbf{89.8}   \\
\multicolumn{2}{|c|}{RPE (PC + MCOM)}   &85.7  &83.0 &87.5  \\
\hline
\end{tabular}
}
\end{table}
\begin{table*}
\setlength{\abovecaptionskip}{0pt}
\setlength{\belowcaptionskip}{5pt}
\centering
\caption{Consistent semantics expressed by relations and corresponding relation paths.}
\begin{tabular}{|c|c|}
\hline
entity pair &(sociology, George Washington University)\\
\hline
relation    &/education/field\_of\_study/students\_majoring./education/education/institution \\
\hline
\multirow{4}{*}{relation paths} &a: /education/field\_of\_study/students\_majoring./education/education/student $\rightarrow$ \\
                                &/people/person/education./education/education/institution \\
                                &b: /people/person/education./education/education/major\_field\_of\_study$^{-1}$ $\rightarrow$ \\
                                &/education/educational\_institution/students\_graduates./education/education/student$^{-1}$ \\
\hline
entity pair &(Planet of the Apes, art director)\\
\hline
relation    &/education/field\_of\_study/students\_majoring./education/education/institution \\
\hline
\multirow{2}{*}{relation paths} &a: /film/film/sequel $\rightarrow$ /film/film\_job/films\_with\_this\_crew\_job./film/film\_crew\_gig/film$^{-1}$\\
                                &b: /film/film/prequel$^{-1}$ $\rightarrow$  /film/film/other\_crew./film/film\_crew\_gig/film\_crew\_role\\
\hline
\end{tabular}

\end{table*}
\subsection{Triple Classification}
We conduct the task of triple classification on three benchmark datasets: FB15K, FB13 and WN11. Triple classification aims to predict whether a given triple (\emph{h,r,t}) is true, which is a binary classification problem.

\subsubsection{Evaluation Protocol}
We set different relation-specific thresholds \{$\delta_{r}$\} to
perform this task. For a test triple (\emph{h,r,t}), if its score
\emph{S}(\emph{h,r,t}) is below $\delta_{r}$, then we predict it as a positive one; otherwise, it is negative. \{$\delta_{r}$\} is obtained by maximizing the classification accuracies on the valid set.
\subsubsection{Implementation}
We directly compare our model with prior work using the results about
knowledge embedding models reported in~\cite{Lin2015LearningEA} for
WN11 and FB13. Because~\cite{Lin2015ModelingRP} does not evaluate
PTransE's performance on this task, we use the code of PTransE that is
released in~\cite{Lin2015ModelingRP} to complete it. FB13 and WN11
already contain negative samples. For FB15K, we use the same process
to produce negative samples, as suggested
by~\cite{Socher2013ReasoningWN}. The hyper-parameter intervals are the
same as link prediction. The best configurations for RPE (PC + ACOM)
are as follows: \emph{n}=50, \emph{m}=50, $\gamma_{1}$=5,
$\gamma_{2}$=6, $\alpha$=0.0001, \emph{B}=1440, $\lambda$=0.8, and $\eta$=0.05, taking the \emph{L$_{1}$} norm on WN11; \emph{n}=100, \emph{m}=100,
$\gamma_{1}$=3, $\gamma_{2}$=6, $\alpha$=0.0001, \emph{B}=960,
$\lambda$=0.8, and $\eta$=0.05, taking the \emph{L$_{1}$} norm on FB13; and
\emph{n}=100, \emph{m}=100, $\gamma_{1}$=4, $\gamma_{2}$=5,
$\alpha$=0.0001, \emph{B}=4800, $\lambda$=1, and $\eta$ =0.05, taking
the \emph{L$_{1}$} norm on FB15K. We exploit RPE (initial) for initiation,
and we set the path length as 2 and the maximum epoch as 500.

\subsubsection{Analysis of Results}
Table 4 lists the results for triple classification on different
datasets, and the evaluation metric is classification accuracy. The
results demonstrate that 1) RPE (PC + ACOM) achieves the best
performance on all datasets, which takes good advantage of
path-specific projection and type constraints; 2) RPE
(PC) improves the performance of RPE (initial) by 4.5\%, 6.0\%, and
13.2\%, particularly on FB15K; thus, we consider that lengthening the
distances for similar entities in embedding space is essential to
specific problems. The results of experiments also indicate that
although LCWA can compensate for the loss for type information, real relation-type information is predominant.
\subsection{Case Study of Consistent Semantics}
As shown in Table 5, for two entity pairs (sociology, George
Washington University) and (Planet of the Apes, art director) from
Freebase, RPE provides two relations and four most relevant relation
paths (each relation is mapped to two relation paths, denoted as a and
b), which are considered as having similar semantics to their respective
relations. However, this type of consistent semantics of reliable
relation paths cannot be achieved by translation-based models, such as
Trans(E, H, R), because translation-based models only exploit the
direct links and do not consider relation path information, such as
reliable relation paths in line 3 and line 6 in Table 5. In contrast,
RPE can obtain reliable relation paths with their consistent
semantics, and it extends the projection and type constraints of the specific relation to the specific path. Furthermore, the experimental results demonstrate that by explicitly using the additional semantics, RPE consistently and significantly outperforms state-of-the-art knowledge embedding models in the two benchmark tasks (link prediction and triple classification).

\section{Conclusions and Future Work}
In this paper, we propose a compositional learning model of relation
path embedding (RPE) for knowledge base completion. To the best of our
knowledge, this is the first time that a path-specific projection has
been proposed, and it simultaneously embeds each entity into relation and path spaces to learn more meaningful embeddings. Moreover, We also put forward the novel path-specific type constraints based on relation-specific constraints to better distinguish similar embeddings in the latent space.
\par
In the future, we plan to 1) incorporate other potential semantic
information into the relation path modeling, such as the information
provided by those intermediate nodes connected by relation paths, and 2) explore relation path embedding in other applications associated with knowledge bases, such as distant supervision for relation extraction and question answering over knowledge bases.
\section*{Acknowledgments}
The authors are grateful for the support of the National Natural
Science Foundation of China (No. 61572228, No. 61472158, No. 61300147,
and No. 61602207), the Science Technology Development Project from
Jilin Province (No. 20140520070JH and No. 20160101247JC), the
Premier-Discipline Enhancement Scheme supported by Zhuhai Government
and Premier Key-Discipline Enhancement Scheme supported by Guangdong
Government Funds, and the Smart Society Collaborative Project funded by the European Commission's 7th Framework ICT Programme for Research and Technological Development under  Grant Agreement No. 600854.

\bibliographystyle{named}
\bibliography{ijcai17}

\begin{thebibliography}{}

\bibitem[\protect\citeauthoryear{Ahn \bgroup \em et al.\egroup
  }{2016}]{Ahn2016ANK}
Sungjin Ahn, Heeyoul Choi, Tanel P{\"a}rnamaa, and Yoshua Bengio.
\newblock A neural knowledge language model.
\newblock {\em CoRR}, abs/1608.00318, 2016.

\bibitem[\protect\citeauthoryear{Bollacker \bgroup \em et al.\egroup
  }{2008}]{bollacker2008freebase}
Kurt Bollacker, Colin Evans, Praveen Paritosh, Tim Sturge, and Jamie Taylor.
\newblock Freebase: a collaboratively created graph database for structuring
  human knowledge.
\newblock In {\em Proceedings of KDD}, pages 1247--1250, 2008.

\bibitem[\protect\citeauthoryear{Bordes \bgroup \em et al.\egroup
  }{2013}]{Bordes2013TranslatingEF}
Antoine Bordes, Nicolas Usunier, Alberto Garc{\'i}a-Dur{\'a}n, Jason Weston,
  and Oksana Yakhnenko.
\newblock Translating embeddings for modeling multi-relational data.
\newblock In {\em Proceedings of NIPS}, pages 2787--2795, 2013.

\bibitem[\protect\citeauthoryear{Carlson \bgroup \em et al.\egroup
  }{2010}]{Carlson2010TowardAA}
Andrew Carlson, Justin Betteridge, Bryan Kisiel, Burr Settles, Estevam
  Hruschka, and Tom Mitchell.
\newblock Toward an architecture for never-ending language learning.
\newblock In {\em Proceedings of AAAI}, pages 1306--1313, 2010.

\bibitem[\protect\citeauthoryear{Chang \bgroup \em et al.\egroup
  }{2014}]{Chang2014TypedTD}
Kai-Wei Chang, Wen tau Yih, Bishan Yang, and Christopher Meek.
\newblock Typed tensor decomposition of knowledge bases for relation
  extraction.
\newblock In {\em Proceedings of EMNLP}, pages 1568--1579, 2014.

\bibitem[\protect\citeauthoryear{Dong \bgroup \em et al.\egroup
  }{2014}]{Dong2014KnowledgeVA}
Xin Dong, Evgeniy Gabrilovich, Geremy Heitz, Wilko Horn, Ni~Lao, Kevin Murphy,
  Thomas Strohmann, Shaohua Sun, and Wei Zhang.
\newblock Knowledge vault: a web-scale approach to probabilistic knowledge
  fusion.
\newblock In {\em Proceedings of KDD}, pages 601--610, 2014.

\bibitem[\protect\citeauthoryear{Dong \bgroup \em et al.\egroup
  }{2015}]{Dong2015QuestionAO}
Li~Dong, Furu Wei, Ming Zhou, and Ke~Xu.
\newblock Question answering over freebase with multi-column convolutional
  neural networks.
\newblock In {\em Proceedings of ACL}, pages 260--269, 2015.

\bibitem[\protect\citeauthoryear{Gardner and
  Mitchell}{2015}]{Gardner2015EfficientAE}
Matthew Gardner and Tom Mitchell.
\newblock Efficient and expressive knowledge base completion using subgraph
  feature extraction.
\newblock In {\em Proceedings of EMNLP}, pages 1488--1498, 2015.

\bibitem[\protect\citeauthoryear{Guu \bgroup \em et al.\egroup
  }{2015}]{Guu2015TraversingKG}
Kelvin Guu, John Miller, and Percy Liang.
\newblock Traversing knowledge graphs in vector space.
\newblock In {\em Proceedings of EMNLP}, pages 318--327, 2015.

\bibitem[\protect\citeauthoryear{Ji \bgroup \em et al.\egroup
  }{2016}]{Ji2016KnowledgeGC}
Guoliang Ji, Kang Liu, Shizhu He, and Jun Zhao.
\newblock Knowledge graph completion with adaptive sparse transfer matrix.
\newblock In {\em Proceedings of AAAI}, pages 985--991, 2016.

\bibitem[\protect\citeauthoryear{Krompass \bgroup \em et al.\egroup
  }{2015}]{Krompass2015TypeConstrainedRL}
Denis Krompass, Stephan Baier, and Volker Tresp.
\newblock Type-constrained representation learning in knowledge graphs.
\newblock In {\em Proceedings of ISWC}, pages 640--655, 2015.

\bibitem[\protect\citeauthoryear{Lao \bgroup \em et al.\egroup
  }{2011}]{Lao2011RandomWI}
Ni~Lao, Tom Mitchell, and William~W. Cohen.
\newblock Random walk inference and learning in a large scale knowledge base.
\newblock In {\em Proceedings of EMNLP}, pages 529--539, 2011.

\bibitem[\protect\citeauthoryear{Lao \bgroup \em et al.\egroup
  }{2015}]{Lao2015LearningRF}
Ni~Lao, Einat Minkov, and William~W. Cohen.
\newblock Learning relational features with backward random walks.
\newblock In {\em Proceedings of ACL}, pages 666--675, 2015.

\bibitem[\protect\citeauthoryear{Lin \bgroup \em et al.\egroup
  }{2015a}]{Lin2015ModelingRP}
Yankai Lin, Zhiyuan Liu, HuanBo Luan, Maosong Sun, Siwei Rao, and Song Liu.
\newblock Modeling relation paths for representation learning of knowledge
  bases.
\newblock In {\em Proceedings of EMNLP}, pages 705--714, 2015.

\bibitem[\protect\citeauthoryear{Lin \bgroup \em et al.\egroup
  }{2015b}]{Lin2015LearningEA}
Yankai Lin, Zhiyuan Liu, Maosong Sun, Yang Liu, and Xuan Zhu.
\newblock Learning entity and relation embeddings for knowledge graph
  completion.
\newblock In {\em AAAI}, pages 2181--2187, 2015.

\bibitem[\protect\citeauthoryear{Miller}{1995}]{Miller1992WORDNETAL}
George~A. Miller.
\newblock Wordnet: a lexical database for english.
\newblock {\em Communications of the ACM}, 38:39--41, 1995.

\bibitem[\protect\citeauthoryear{Neelakantan \bgroup \em et al.\egroup
  }{2015}]{Neelakantan2015CompositionalVS}
Arvind Neelakantan, Benjamin Roth, and Andrew McCallum.
\newblock Compositional vector space models for knowledge base completion.
\newblock In {\em Proceedings of ACL}, pages 156--166, 2015.

\bibitem[\protect\citeauthoryear{Nickel \bgroup \em et al.\egroup
  }{2016}]{Nickel2016ARO}
Maximilian Nickel, Kevin Murphy, Volker Tresp, and Evgeniy Gabrilovich.
\newblock A review of relational machine learning for knowledge graphs.
\newblock {\em Proceedings of the IEEE}, 104:11--33, 2016.

\bibitem[\protect\citeauthoryear{Richardson and
  Domingos}{2006}]{Richardson2006MarkovLN}
Matthew Richardson and Pedro~M. Domingos.
\newblock Markov logic networks.
\newblock {\em Machine Learning}, 62:107--136, 2006.

\bibitem[\protect\citeauthoryear{Riedel \bgroup \em et al.\egroup
  }{2013}]{Riedel2013RelationEW}
Sebastian Riedel, Limin Yao, Andrew McCallum, and Benjamin~M. Marlin.
\newblock Relation extraction with matrix factorization and universal schemas.
\newblock In {\em Proceedings of HLT-NAACL}, pages 74--84, 2013.

\bibitem[\protect\citeauthoryear{Socher \bgroup \em et al.\egroup
  }{2013}]{Socher2013ReasoningWN}
Richard Socher, Danqi Chen, Christopher~D. Manning, and Andrew~Y. Ng.
\newblock Reasoning with neural tensor networks for knowledge base completion.
\newblock In {\em Proceedings of NIPS}, pages 926--934, 2013.

\bibitem[\protect\citeauthoryear{Suchanek \bgroup \em et al.\egroup
  }{2007}]{suchanek2007yago}
Fabian~M. Suchanek, Gjergji Kasneci, and Gerhard Weikum.
\newblock Yago: a core of semantic knowledge.
\newblock In {\em Proceedings of WWW}, pages 697--706, 2007.

\bibitem[\protect\citeauthoryear{Toutanova \bgroup \em et al.\egroup
  }{2016}]{Toutanova2016CompositionalLO}
Kristina Toutanova, Victoria Lin, Wen tau Yih, Hoifung Poon, and Chris Quirk.
\newblock Compositional learning of embeddings for relation paths in knowledge
  base and text.
\newblock In {\em Proceedings of ACL}, pages 1434--1444, 2016.

\bibitem[\protect\citeauthoryear{Wang \bgroup \em et al.\egroup
  }{2014}]{Wang2014KnowledgeGE}
Zhen Wang, Jianwen Zhang, Jianlin Feng, and Zheng Chen.
\newblock Knowledge graph embedding by translating on hyperplanes.
\newblock In {\em Proceedings of AAAI}, pages 1112--1119, 2014.

\bibitem[\protect\citeauthoryear{Wang \bgroup \em et al.\egroup
  }{2015}]{Wang2015KnowledgeBC}
Quan Wang, Bin Wang, and Li~Guo.
\newblock Knowledge base completion using embeddings and rules.
\newblock In {\em Proceedings of IJCAI}, pages 1859--1865, 2015.

\bibitem[\protect\citeauthoryear{Wang \bgroup \em et al.\egroup
  }{2016}]{Wang2016KnowledgeBC}
Quan Wang, Jing Liu, Yuanfei Luo, Bin Wang, and Chin-Yew Lin.
\newblock Knowledge base completion via coupled path ranking.
\newblock In {\em Proceedings of ACL}, pages 1308--1318, 2016.

\bibitem[\protect\citeauthoryear{Xie \bgroup \em et al.\egroup
  }{2016}]{Xie2016RepresentationLO}
Ruobing Xie, Zhiyuan Liu, and Maosong Sun.
\newblock Representation learning of knowledge graphs with hierarchical types.
\newblock In {\em Proceedings of IJCAI}, 2016.

\end{thebibliography}

\end{document}